\documentclass[letterpaper, 10 pt, conference]{ieeeconf}  

\IEEEoverridecommandlockouts                              
\usepackage{graphics} 
\usepackage{epsfig} 
\usepackage{mathptmx} 
\usepackage{times} 
\usepackage{amsmath} 
\usepackage{amssymb}  
\usepackage{mathtools}
\usepackage{multirow}
\usepackage{color}
\usepackage{hyperref}
\let\labelindent\relax
\usepackage{enumitem}
  
\hypersetup{colorlinks, linkcolor=red}

\title{\LARGE \bf
Maintaining Grasps within Slipping Bound by Monitoring Incipient Slip 
}

\author{
  \authorblockN{Siyuan Dong, Daolin Ma, Elliott Donlon, Alberto Rodriguez} 
  \authorblockA{
     Massachusetts Institute of Technology\\
    {\tt\small <sydong,daolinma,edonlon,albertor>@mit.edu}} 
\thanks{This work was supported by the Amazon Research Awards, and the Toyota Research Institute (TRI). 
This article solely reflects the opinions and conclusions of its authors and not Amazon or Toyota.}}

\begin{document}

\maketitle
\thispagestyle{empty}
\pagestyle{empty}

\begin{abstract}
In this paper, we propose an approach to detect incipient slip, i.e. predict slip, by using a high-resolution vision-based tactile sensor, GelSlim. The sensor dynamically captures the tactile imprints of the contact object and their changes with a soft gel pad. The method assumes the object is mostly rigid and treats the motion of object's imprint on sensor surface as a 2D rigid-body motion. We use the deviation of the true motion field from that of a 2D planar rigid transformation as a measure of slip. The output is a dense slip field which we use to detect when small areas of the contact patch start to slip (incipient slip). The method can detect both translational and rotational incipient slip without any prior knowledge of the object at 24 Hz. We test the method on 10 objects 240 times and achieve 86.25\% detection accuracy. We further show how the slip feedback can be used to monitor the gripping force to avoid slip with a closed-loop bottle-cap screwing and unscrewing experiment with incipient slip detection feedback. The method was demonstrated to be useful for the robot to apply proper gripping force and stop screwing at the right point before breaking objects. The method can be applied to many manipulation tasks in both structured and unstructured environments. 

\end{abstract}


\section{INTRODUCTION}
Human hands can perform complex manipulations because of the various tactile sensors distributed in the skin. 
One important function of these tactile sensors is to detect incipient slip. People detect incipient slip by sensing the stretch of the skin~\cite{vallbo1984properties}, which enables them to naturally control forceful manipulation tasks such as adjusting grasping force, resisting external perturbances, following the contours of objects or using tools while maintaining firm grasps. 

\begin{figure}[t]
	\centering
	\includegraphics[width=0.96\linewidth]{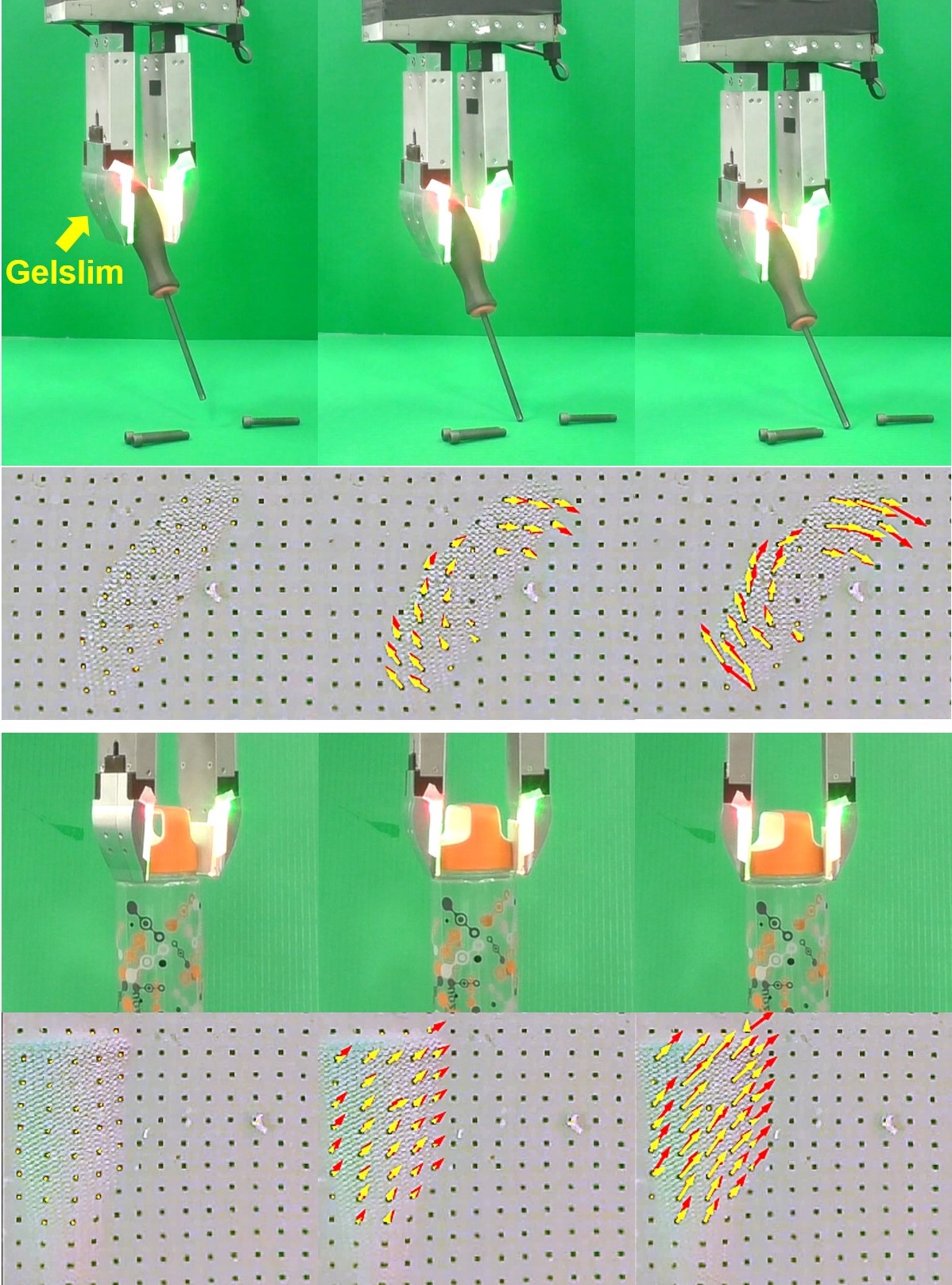}
    \vspace{-4pt}
	\caption{Top: the robot arm is dropping a tilted screw-driver to the ground such that the screw driver slips rotationally when it collides with the ground. Bottom: the robot is screwing a bottle cap, and translational slip starts when the cap is screwed to the end of the threads. The second and fourth rows show the evolution of processed GelSlim images from no shear or slip, to shear with no slip, to full slip for each scenario. The arrows represent real marker displacements (in yellow) and estimated marker displacements (in red) under assumption of no slippage.}
	\label{fig:figure1}
\end{figure}

One major limitation of current robotic systems is their inability to control gripping force -- resulting largely from the lack of tactile sensors capable of providing similar incipient slip detection feedback to that of human skin. Slip detection helps to minimize wear on robotic fingers while also maintaining a firm grasp. Closing the loop with slip detection could aid in tasks such as: a water bottle slipping out of a grasp due to moving liquid, a tool dropping from the hand as a consequence of exerting too much force, or a broken object due to excessive grasping force. 

In this paper, we propose a general purpose incipient slip detection method using the GelSlim tactile sensor~\cite{GelSlim_v1} to enable more intelligent slip-aware grasp control. As shown in Figure~\ref{fig:figure1} top row, the robot can sense rotational slippage when the screw driver collides with the ground and then naturally decreases the grasping force to gently place it on the ground. As shown in Figure~\ref{fig:figure1} bottom row, when screwing the bottle cap, the robot can also determine the correct stopping point by detecting translational slip in the process.

The GelSlim sensor~\cite{GelSlim_v1}, shown in Figure~\ref{fig:figure1}, is a vision-based tactile sensor. It uses a Raspberry Pi Spy Camera to capture a deformation of a gel-based skin on the sensor surface to measure the contact position and local texture of the object. The gel is covered by a piece of fabric, which functions as a protective layer and enhances the contact signal with its own texture. The tactile imprints of the screw driver handle and bottle cap captured by GelSlim sensor are shown in the second and fourth row (respectively) of Figure~\ref{fig:figure1}. On the surface of the gel, there are evenly distributed black markers. The markers move under contact force and the displacement field of the markers represents how much the gel is stretched in and around the contact region. 

We define incipient slip as the condition under which parts of the contact region (generally the peripheral) start to slip. The marker displacement in slip region must be smaller than the displacement of the object. The real marker displacement field in the contact region is shown with yellow arrows in Figure~\ref{fig:figure1}. Assuming no slip happens in the central region of the contact patch, we model the movement of the object-gel contact region as a 2D rigid body to calculate the estimated marker displacement field (shown with red arrows in Figure~\ref{fig:figure1}). If the real and estimated displacement fields are similar, no slip happens (see second column of the GelSlim images in Figure~\ref{fig:figure1}). However, if they have quite amount of difference, incipient slip happens (see third column of the GelSlim images). Based on the principle described above, our incipient slip detector can detect both translational and rotational slippage without any prior knowledge of the object, such as coefficient of friction, mass or shape. With a light computation cost, the detector can run at around 24 Hz.

We test the detection accuracy of the method with 10 daily objects under a large number of external forces, and achieve 86.25\% success rate. We also successfully implement closed-loop bottle cap screwing and unscrewing experiments with incipient slip detection feedback, which informs the robot when to increase the gripping force and when to stop. The method can also be applied to other vision-based tactile sensors with dense markers on the surface~\cite{GelSight_Dong, GelForce2005} and it can find applications in manipulation tasks like: in-hand manipulation~\cite{yousef2011tactile}, closed-loop grasp control~\cite{Zeng2018}, contour-following~\cite{hellman2018functional}.


\section{RELATED WORK}


Various tactile sensors have been developed in the past few decades specially to detect incipient slip or slip~\cite{slipreview2013}. There are mainly two categories of working principles: vibration based and relative motion based slip detection.

\textbf{Vibration based method} Since relative motion between two rough surfaces results in high-frequency signals, many tactile sensors have been built to detect slip by sensing vibration~\cite{veiga2015stabilizing,li2014learning}. Sch{\"o}pfer~\textit{et al}.~\cite{schopfer2010_piezo} proposed a incipient slip detection method by using piezo-resistive tactile sensor, which could measure the normal force at 83 Hz within 80 mm $\times$ 80 mm sensing area. The time-sequence force data was reprocessed by using a Fourier transform first and then fed into a neural network to detect slip. Sch{\"u}rmann~\textit{et al}.~\cite{schurmann2012highspeed} further improved the frame rate up to 1.9k Hz and successfully implemented grasp force control with slip detection feedback. The method is able to detect slip with unknown objects, however, the configuration of the tactile sensor is hard to directly to grasp objects with. Romano~\textit{et al}.~\cite{romano2011human} used the vibration signals from an accelerometer (running at 3 kHz) and pressure cells (running at 24.4 Hz) embedded in the palm of the gripper to detect slip. They also performed a similar grasp force control experiment with slip feedback control at around 1K Hz with a PR2 robot. 

\textbf{Relative motion based method} Relative motion based slip detection is mostly utilized in vision-based tactile sensors~\cite{ueda2005development,ueda2005grip}. Maldonado~\textit{et al}.~\cite{maldonado2012improving} designed a compact fingertip sensor, where a small camera with 30 $\times$ 30 pixels spatial resolution and 1 $mm^2$ sensing region is embedded inside, to detect the relative motion between the object and fingertip. Given the area of the sensor's contact region, it is hard for this sensor to detect slip when the object has no texture. The GelSight sensor~\cite{GelSightUSB, GelSight_Dong}, which is also a vision-based sensor but uses an camera to capture the deformation and stretch of an elastomer gel, was also used to detect slip. Yuan~\textit{et al}.~\cite{GelSightShear} proposed a method to detect incipient slip by monitoring the entropy of the marker displacement according to the effect that the stretch of the gel surface is inhomogeneous in slip state. However, this method only works well when the contact surface has little texture. Dong~\textit{et al}.~\cite{GelSight_Dong} used the GelSight sensor to measure the relative motion between the markers on the gel surface, object texture, and maker displacement ratio between the center contact region and the peripheral area, which achieved $71\%$ test accuracy with $37$ different daily objects. These two mechanisms are complementary for objects with \& without textures, however, the relatively small sensing area of the GelSight sensor, and the relatively large computational cost makes it difficult to use in closed-loop control scenarios. Li~\textit{et al}.~\cite{li2018slip} further improved the detection accuracy to 88\% by adding an external camera beside the GelSight sensor and training a convolutional neural network. However, this technique further decreases the manipulation speed. Compared to these methods above, our approach speeds up the detecting frame rate by using a simpler algorithm, and achieves similar/better detection accuracy meanwhile. In addition, compared to other vision-based sensors, the GelSlim sensor has much larger sensing region ($3cm \times 4cm$), which also provides more buffer time to sense and react to the slip event.

\section{Method} \label{method}


\textbf{Physical principle} When a small portion of the contact area has lost contact, we call this condition ``incipient slip". Under a shear load, the gel surface will be stretched along the force direction. This stretch is easily calculated by tracking the markers on the gel surface. As the force increases, gel in the contact area will move with the object until the frictional force is not sufficient to hold the contact. Before incipient slip happens, the gel in the contact region has the same displacement as the object. When slip starts, the peripheral edge of the contact region will slip first~\cite{andre2011effect}, which means the displacement field of that region will have smaller magnitude than that at the center. This is the trigger signal for our method to detect incipient slip.


\textbf{Algebra} The method starts from the core assumption that under no slip, the contact patch will move as a planar rigid body. The motion of every point in the contact patch can then be considered as pure rotation around a point, which is called the instantaneous center of rotation (ICR). If we know the position ($x$, $y$), velocity ($\upsilon_{x}$, $\upsilon_{y}$) and angular velocity $\omega$ of a reference point in the rigid body, we can calculate the velocity ($\upsilon_{x}$, $\upsilon_{y}$) and angular velocity $\omega'$ of any other arbitrary point with known position ($x'$, $y'$) inside the rigid body according to equation~\ref{eq:1}. If the point ($x$, $y$) is the ICR, $\upsilon_{x}$ and $\upsilon_{y}$ are zero. 

\begin{equation}
  \label{eq:1}
  \begin{gathered}
    \upsilon_{x'} = \upsilon_{x} + \omega*(y-y')\\        
    \upsilon_{y'} = \upsilon_{y} + \omega*(x'-x)\\
    \omega' = \omega
  \end{gathered}
\end{equation}

As explained above, generally the peripheral region with lower pressure distribution loses contact first, but the inner part of the contact region still holds when incipient slip happens. So we can use the markers in the inner contact region as the reference points of the rigid body motion to calculate the estimated displacement field of the markers in the contact region. When the real displacements of several points in the contact region are sufficiently different from the estimated displacements, we trigger incipient slip.

\textbf{Algorithm} The algorithm includes 4 steps (illustrated in Figure~\ref{fig:steps}). We use a translational incipient slip case as an example (Figure~\ref{fig:steps} left column) to explain in detail. 
\begin{itemize}[leftmargin=*]
\item \textbf{Step1: Detect contact region} since the contact region is highlighted by the fabric textures (Figure~\ref{fig:steps}(a1)), we use the Canny filter to detect the edges and several morphological filters to group the edges together resulting in the contact region (highlighted with green color in Figure~\ref{fig:steps}(b1)). 

\begin{figure}[h]
	\centering
	\includegraphics[width=0.95\linewidth]{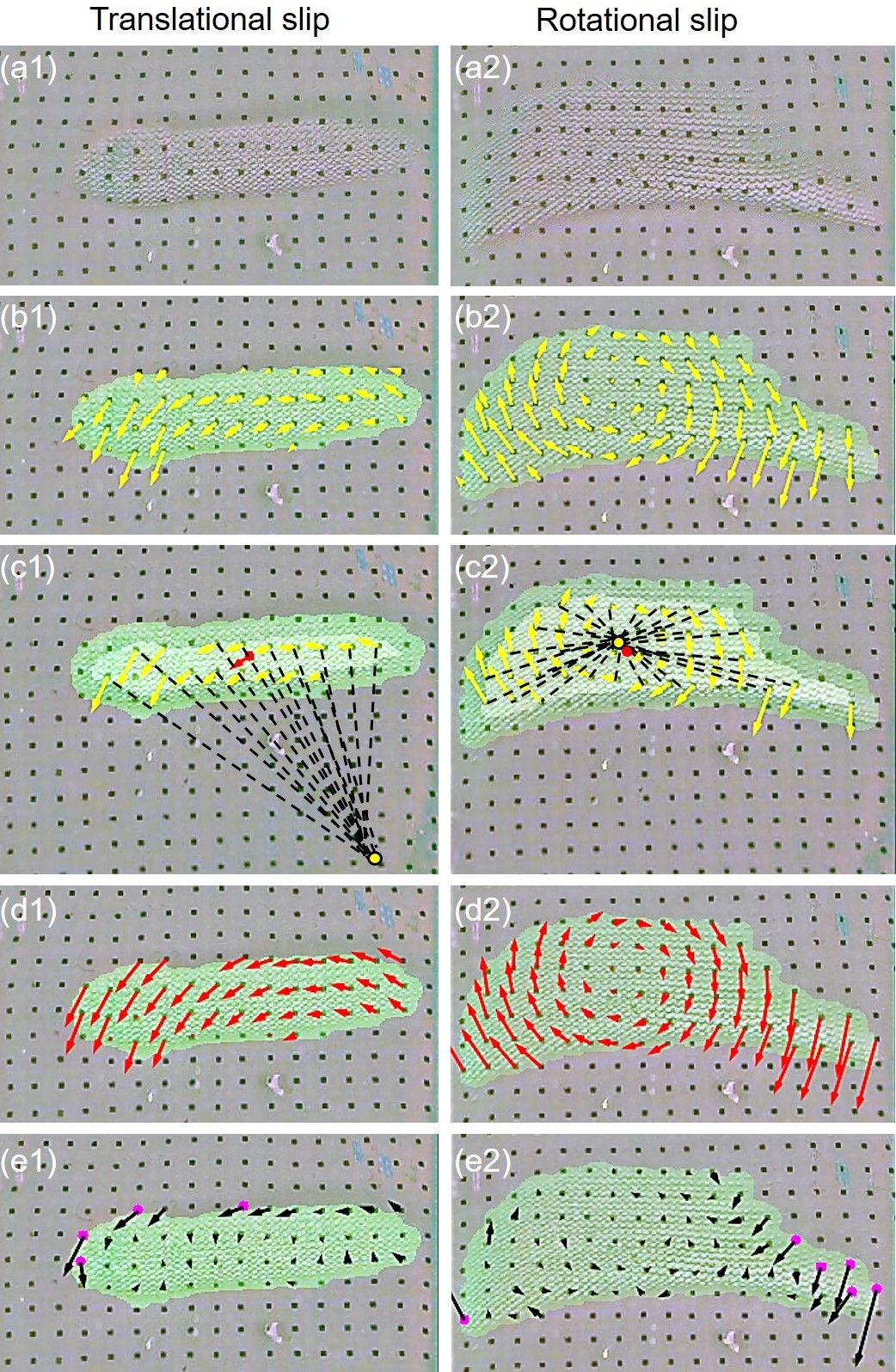}
    \vspace{-4pt}
	\caption{The procedure to detect incipient translational slip (left column) and rotational slip (right column).}
	\label{fig:steps}
\end{figure}

\item \textbf{Step2: Calculate real displacement field} Then the black markers are selected with a threshold and we deploy the SimpleBlobDetector function in OpenCV to detect the position of each marker in the contact region. Through matching and comparing the relative displacement of the marker positions in the reference frame and current frame, we compute the displacement of each selected marker according to the reference frame. The displacement vectors are shown in Figure~\ref{fig:steps}(b1) with yellow arrows. The reference frame we use is captured in the beginning. However, after incipient slip signal is detected, we update the reference frame with the current frame (i.e.,update contact region and reference marker positions) to make the method work dynamically. 

\item \textbf{Step3: Estimate rigid motion of reference point} The next step is to find the inner region and localize the markers inside to find the reference point of the rigid body motion. A simple erosion filter is used to remove the edge region of the contact area and the markers in the remaining area (highlighted with white in Figure~\ref{fig:steps}(c)) are used to calculate the position $(x, y)$, velocity ($\upsilon$) and angular velocity ($\omega$) of the reference point. Since the positions and displacement of these markers are already computed from the last step, we feed them into the estimateRigidTransform function in OpenCV to compute the rigid transform matrix. The angular velocity of the reference point can be extracted from the transform matrix, and the position and velocity are computed by averaging the positions and displacements of the markers inside. For small motion, we use marker displacements as velocities. The reference point and its velocity (displacement) are labeled with a red arrow in Figure~\ref{fig:steps}(c1). We labeled the ICR with a yellow dot in Figure~\ref{fig:steps}(c1). The black dashed lines are perpendicular to their corresponding displacements vectors. 

\item \textbf{Step4: Calculate estimated displacement field and slip field} we calculate the estimated marker displacement filed in the contact region (shown with red arrows in Figure~\ref{fig:steps}(d1)) according to Equation~\ref{eq:1}. The slip field are the difference between the real and estimated marker displacement fields (black arrows in Figure~\ref{fig:steps}(e1)). If multiple markers (labeled with purple dots) in the slip field are larger than a threshold, it indicates that incipient slip has happened. We notice that all of them lie in the peripheral region, which also verifies the phenomenon described in~\cite{andre2011effect}. 
\end{itemize}

\begin{figure}[h]
	\centering
	\includegraphics[width=0.8\linewidth]{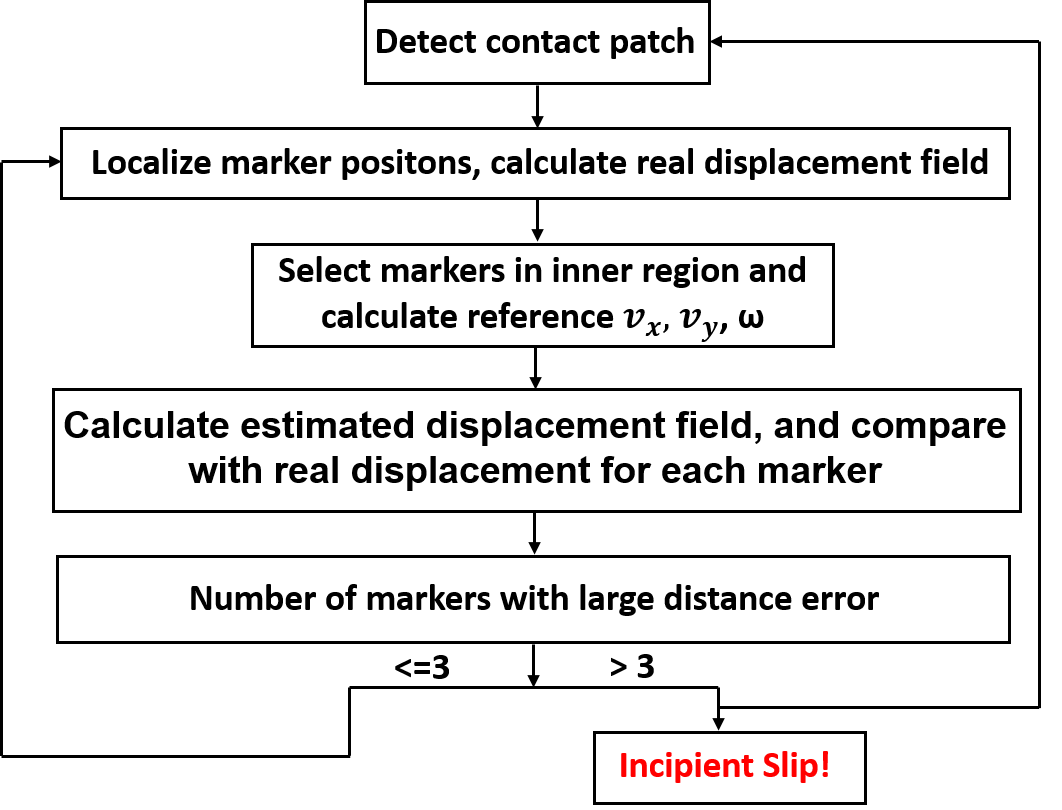}
    \vspace{-4pt}
	\caption{Incipient slip detection algorithm pipeline}
	\label{fig:pipeline}
\end{figure}

The right column of Figure~\ref{fig:steps} shows the same procedure on detecting rotational slip. Since the object rotates in the clockwise direction and the ICR is roughly in the center of the contact patch, the upper right and down left region lose contact first, which can be clearly seen from the slip field. To make the algorithm clearer, we include Figure~\ref{fig:pipeline} to show the algorithm pipeline.

This method can detect any type of directional slip (translational, rotational or both) as long as the contact patch spans a cluster of markers. The rotation center for rotational slip can be easily computed if needed. It does not require any prior knowledge of the object and contact, such as coefficient of friction, mass or shape. The method can run at around 24 Hz and the largest computational cost is due to on localizing the marker positions, which means smaller contact patch results in a higher speed. In addition, we can expect large improvement from parallel or GPU computation.

\section{Experiment}

\subsection{Experiment Setup} 
We conduct experiments in a setup based on the grasping system developed in~\cite{Zeng2018,hogan2018tactile} with a 6-DOF ABB-1600 robot arm equipped with a WSG-50 parallel gripper. Two GelSlim sensors are mounted in the gripper as two fingers (Figure~\ref{fig:figure1}). The GelSlim sensor captures the imprints of the contact surface with 640*480 spatial resolution at 30 Hz. The sensing surface of the sensor measures $3\times4$ $cm^2$. In this experiment, for computational simplicity, only one of the two sensors is used to detect incipient slip. 

\subsection{Incipient Slip Detection Test} 

We test the accuracy of detecting incipient slip on 10 daily objects (Table~\ref{tab:objects}) with different shapes, masses, softness and friction coefficients. The yellow duck, brain foam and scrub sponges are soft and deformable. The water bottle (full of water) is much heavier than other objects. The crayons box, scrub sponges and glue bottle have flat surfaces. The surfaces of scrub sponges, brain foam and glue bottle are smooth to the sensor surface. 

Each object is initially held by the gripper and then pushed, pulled or rotated manually. For translational slip, we pushed the objects along 4 evenly spaced directions in the plane parallel to the finger phalanges. For rotational slip, we rotate the objects in clockwise and anticlockwise directions. For the non-slip case, we try to push or rotate the object without making it slip. Since it is difficult to accurately control the whether or not the object slips in this open-loop experiment, we relabel the data afterwards according to human observation. To test the algorithm in different working scenarios, we repeat the experiment with three different forces (from 5 to 30N for most objects). Therefore, each object is tested 24 times (18 slip cases and 6 non-slip cases) for a total of 240 experiments. The algorithm is running in real time, and the results are saved automatically. 

\begin{table*}[] 
\centering
\caption{Ten daily objects used to test the algorithm and the detection accuracy} \label{tab:objects} 
\begin{tabular}{c|c|c|c|c|c|c|c|c|c|c|c}
\hline
  &  Tape & Crayons &  Scissor & \vtop{\hbox{\strut Water}\hbox{\strut bottle}} & \vtop{\hbox{Super}\hbox{\strut glue}} & \vtop{\hbox{\strut Brain}\hbox{\strut foam}} & \vtop{\hbox{\strut Toy}\hbox{\strut duck}} & \vtop{\hbox{\strut Screw}\hbox{\strut driver}} & \vtop{\hbox{\strut Scrub}\hbox{\strut sponges}} & \vtop{\hbox{\strut Blue}\hbox{\strut tube}} & Total \\ \hline
   & \includegraphics[width=0.06\textwidth]{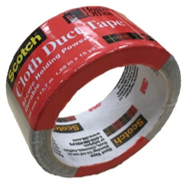}  & \includegraphics[width=0.05\textwidth]{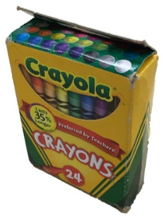}  & \includegraphics[width=0.032\textwidth]{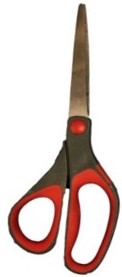}  & \includegraphics[width=0.033\textwidth]{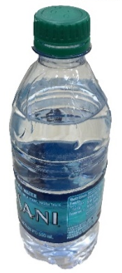}  & \includegraphics[width=0.033\textwidth]{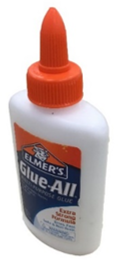}  & \includegraphics[width=0.055\textwidth]{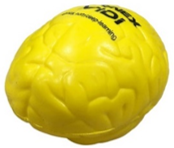}  & \includegraphics[width=0.05\textwidth]{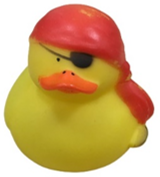}  & \includegraphics[width=0.03\textwidth]{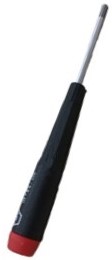}  & \includegraphics[width=0.05\textwidth]{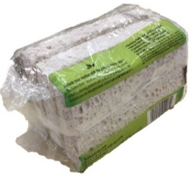} & \includegraphics[width=0.05\textwidth]{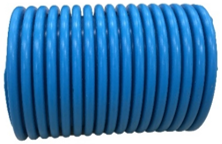}  \\ \hline
 Success Rate & 100.0\%  & 49.97 \% & 100.0\% & 100.0\% & 95.8\% & 87.5\% & 100.0\% & 100\% & 33.33\% & 95.83\% & \color{blue}{86.25\%} \\ \hline 
 Failure (False Positives) & 0.0\% & 4.17\% & 0.0\% & 0.0\% & 4.17\% & 0.0\% & 0.0\% & 0.0\% & 0.0\% & 0.0\% & 0.83\% \\ \hline 
 Failure (False Negatives) & 0.0\% & \color{red}{45.83\%} & 0.0\% & 0.0\% & 0.0\% & 12.5\% & 0.0\% & 0.0\% &  \color{red}{66.67\%} & 4.17\% & 12.92\% \\ \hline 
\end{tabular}
\end{table*}


\subsection{Bottle Cap Screwing and unscrewing with Incipient-Slip Feedback}
\begin{figure}[t]
	\centering
	\includegraphics[width=0.9\linewidth]{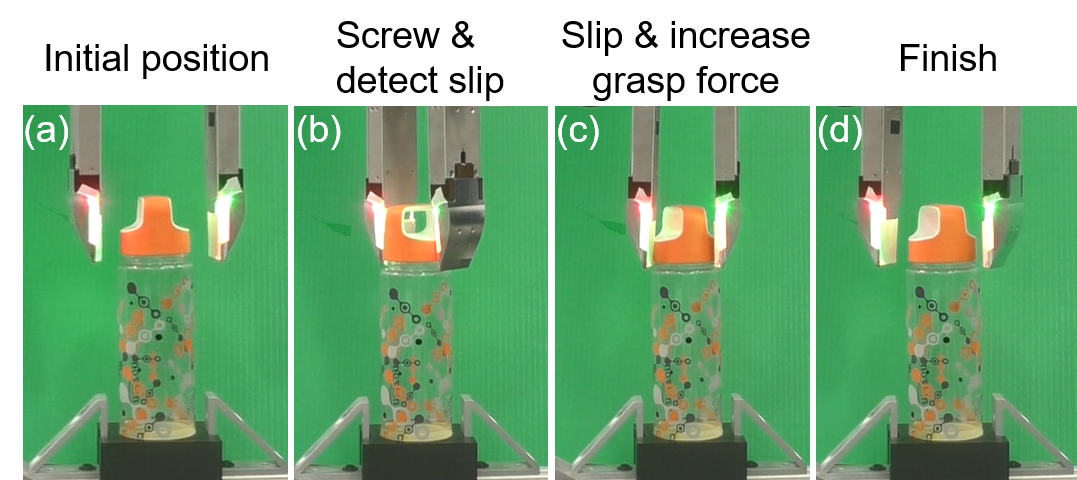}
     \vspace{-4pt}
	\caption{The experimental setup and steps for screwing process}
	\label{fig:screw_experiment} 
\end{figure}
There are many manipulation tasks in daily life that require exerting force or torque while maintaining stable grasps. We propose using incipient-slip detection to implement them. To demonstrate the function of our incipient slip detector, we performed bottle cap screwing and unscrewing experiments, where the detector informs when the robot increases the gripping force and stops screwing. 

The experimental setup and screwing process are shown in Figure~\ref{fig:screw_experiment}. The water bottle is fixed on a platform, and the cap is initially placed on top of the bottle loosely. We assume the  position of the bottle cap is known. The gripper is aligned with the central axis of the bottle, and the sensing region of the GelSlim sensor is horizontally aligned with the bottle cap (Figure~\ref{fig:screw_experiment}(a)). The robot then grips the cap with 10N as the initial force and starts with a low speed screwing motion. Because of the compliance of the gel, we do not need to know the exact pitch of the thread. While the robot screws the cap, our incipient slip detector gives warnings (Figure~\ref{fig:screw_experiment}(b)). When incipient slip is detected, we stop the robot at that point momentarily, increase the gripping force by 10N and start the screwing again (Figure~\ref{fig:screw_experiment}(c)). The process is repeated until the force reaches 60N, when we stop the screwing process and open the gripper (Figure~\ref{fig:screw_experiment}(d)). The data of the GelSlim sensor is recorded on-line for posterior analysis (Section~\ref{Results}).

\begin{figure}[h]
	\centering
	\includegraphics[width=0.9\linewidth]{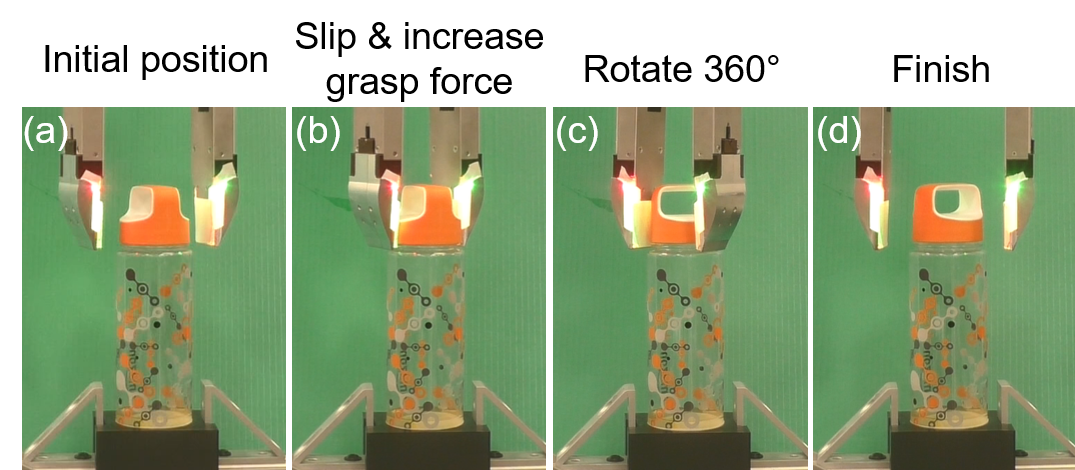}
    \vspace{-4pt}
	\caption{The experiment setup and steps for unscrewing process}
	\label{fig:unscrew_experiment} 
\end{figure}

For the unscrewing experiment, the process is reversed (Figure~\ref{fig:unscrew_experiment}). The cap starts tightly screwed. We still select 10N as the initial force and 60N as the maximum force. The only difference between these two experiments replies on the time when slip happened. For unscrewing process, slip happens at the beginning and it is not necessary for the gripper to increase to the max force limit. This is in contrast to, the screwing experiment where using maximum gripping force allows the robot to firmly torque the bottle cap on. We repeat the two experiments 3 times to test the stability of the algorithm.

\section{Experiment Results and Analysis}\label{Results}
\subsection{Incipient Slip Detection Accuracy}

Table~\ref{tab:objects} summarizes the detection accuracy for each object. Based on the experiment type and the output of the detector, we group the results into three different classes: successful, false positives and false negatives. False positives (0.83\%) means the detector gives false alarm, i.e., the detector is triggered without slip event. False negative (12.90\%) means the detector fails to detect incipient slip when it happens. In general, the detector provides correct detection results for most of the objects with nearly 100\% accuracy except for the crayon box and scrub sponges. On average, the detection accuracy of our method is 86.25\%. 

\begin{figure}[t]
	\centering
	\includegraphics[width=0.8\linewidth]{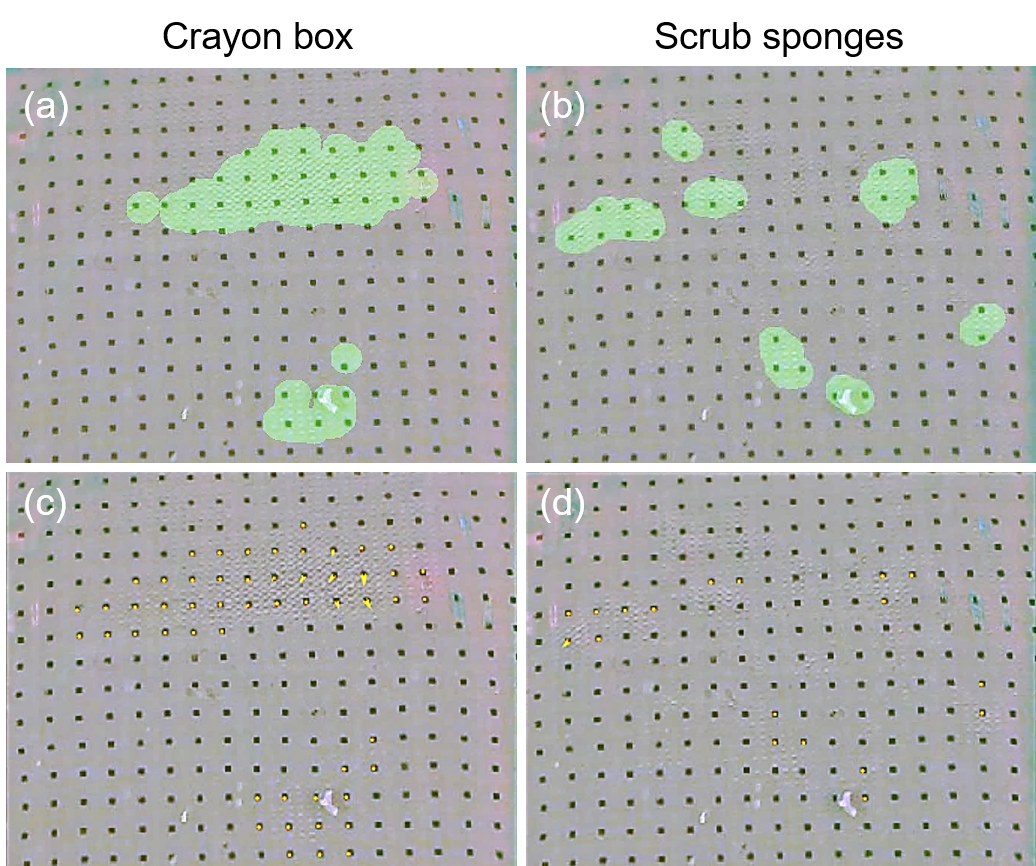}
    \vspace{-4pt}
	\caption{The contact regions of the crayon box (a) and the scrub sponges (b) are highlighted with a green mask. The real marker displacements (in yellow) and estimated marker displacements (in red) are shown when the crayon box (c) and scrub sponges (d) slip}.
	\label{fig:failure} 
\end{figure}

Most of the false negatives errors (12.92\%) result from the experiments with the crayon box and scrub sponges. A more detailed analysis indicates that the contact signals for these two objects are too weak for the algorithm to work (see Figure~\ref{fig:failure}). The two objects with flat and smooth surface make little pressure and shear force on the sensor surface under small gripping force.  The shallow texture and tiny displacement field shown in Figure~\ref{fig:failure} corroborate this point. Therefore, for objects with smooth and flat surfaces, we need to apply larger grasping force to generate a good contact signal for our method to work. We redid the same experiment for these two objects with 50N gripper force. The success rate increased to 100.0\% for the crayon box and 66.67\% for the scrub sponges. It is interesting to note since the quality and size of the contact patch can be measured in real time and used as an assessment if the algorithm is reliable.

\subsection{Bottle Cap Screwing}
\begin{figure*}[t]
	\centering
	\includegraphics[width=0.9\linewidth]{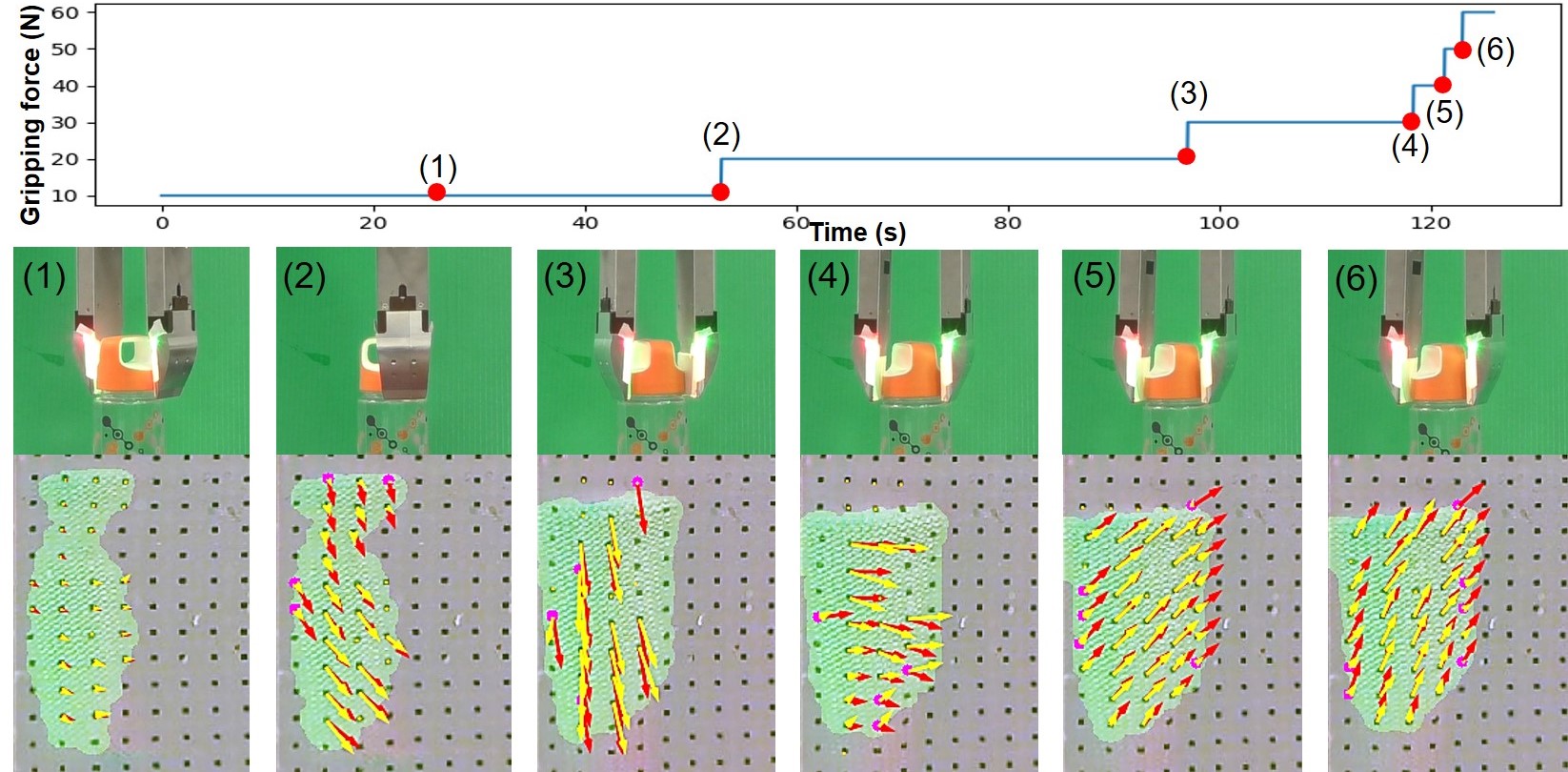}
	\caption{The force change over time (top), the images taken with external camera (middle) and the GelSlim images (bottom) at the time spots labelled in the plot for the bottle cap screwing process.}
	\label{fig:exp1} 
\end{figure*}
In the bottle cap screwing experiment, the incipient slip detector is helpful to determine when increased gripping force is needed and when the cap is already tightly screwed. 
The robot was able to tightly screw the bottle cap for the three experiment trials. We recorded the gripping force change over time, the GelSlim images and external camera images at each detected slip point, which are visualized in Figure~\ref{fig:exp1}. For the GelSlim images, we still use yellow arrows to represent real marker displacements, red arrows to represent estimated marker displacements and the slipped markers are labeled in purple. 

In the first half of the process, the bottle cap was loose and the gel surface shown in Figure~\ref{fig:exp1}(1) was barely stretched. The red and yellow arrows nearly overlapped with each other. The first slip signal was triggered in the middle of the process, the small error in z axis made the arrows in Figure~\ref{fig:exp1}(2) point down. After the gripping force increased to 30N, the contact patch in time (3) also expanded. As the the cap became screwed on more, the accumulated error in z axis triggered another slip, which is shown in Figure~\ref{fig:exp1}(3). Since the cap was already in the thread, the slip actually helped to correct the error. The last three detected slippage happened closely in the last few seconds, which indicates that the cap was almost at the end of the thread. The large maker displacements in the three GelSlim images indicates that the gripper was exerting larger forces on the cap. Since the reference frame is updated right after incipient slip is detected, the displacements in the next frame will be very small. However, after we cleared the displacements in time (4) and (5), the marker displacements in time (5) and (6) were still large, which implies that slip happened continuously in this time period and was continuously detected. The unchanged bottle cap orientation in external images (5) and (6) also validate this point. It is interesting to note that the directionality of the slip field could potentially be used to distinguish between vertical slip (cap goes down, OK) and horizontal slip (lose the grasp, NOT OK).

\subsection{Bottle Cap Unscrewing}
\begin{figure}[h]
	\centering
	\includegraphics[width=0.9\linewidth]{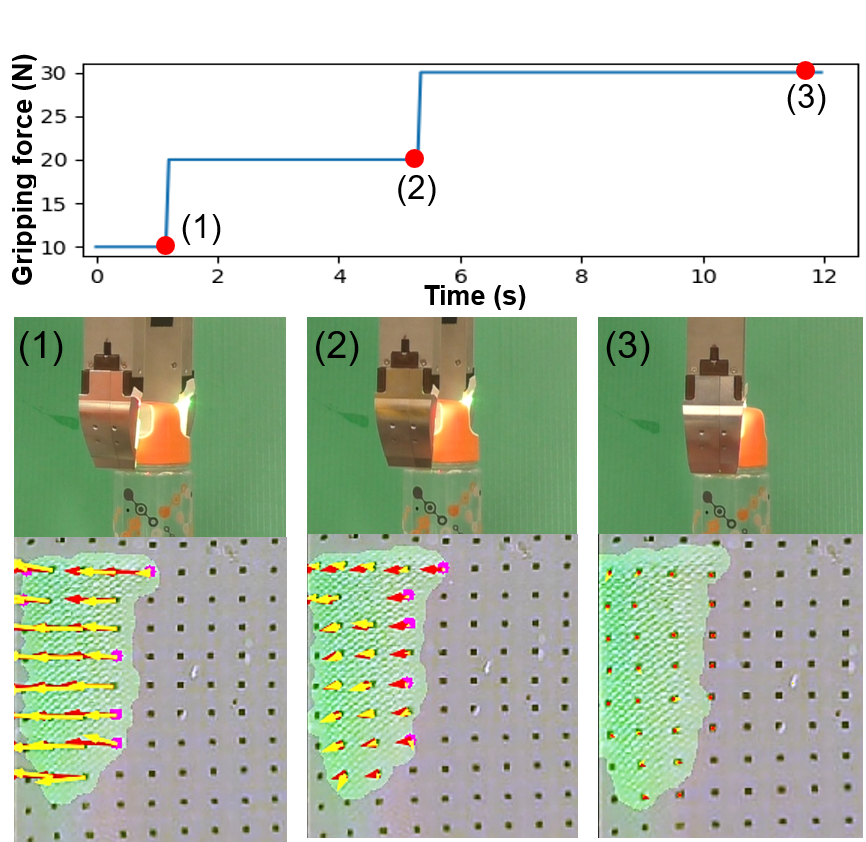}
	\caption{The force change over time (top), the images taken with external camera (middle) and the GelSlim images (bottom) at the time points labelled in the plot for the bottle cap unscrewing process.}
	\label{fig:unscrew} 
\end{figure}
For the bottle cap unscrewing process, incipient slip happened right after we started the experiment, which can be seen in Figure~\ref{fig:unscrew}(1). The arrows point to the opposite direction compared to those in screwing process, since the two tasks have opposite force directions. The second slip was triggered closely to the first one, which implies the cap did not move with the sensor yet. The cap in the external images of Figure~\ref{fig:unscrew}(1)\&(2) also maintains the same orientation. Right after the gripping force increased to 30N, the bottle cap started to follow the gripper, which can be seen by comparing the orientations of the bottle cap in Figure~\ref{fig:unscrew}(2)\&(3). Afterwards, there was no more slip event happened and the marker displacements kept with small values (Figure~\ref{fig:unscrew}(3)). It is worth nothing that all of the slip happens for markers located in peripheral region of the contact area. The bottle cap was successfully opened in all of the three trials.

\section{CONCLUSIONS}

We have proposed an incipient slip algorithm that exploits the displacement field tracking capability of the GelSlim tactile sensor~\cite{GelSlim_v1}. The key idea is to compare the displacement field in the contact patch with that of a perfectly rigid-body displacement. The algorithm achieves 86.25\% detection accuracy in the 240 slip experimental trials with 10 daily objects, which is comparable to~\cite{li2018slip} but at least 8 times faster (24 Hz for our method). This method works best for the objects that make large contact patches and apply large tangential forces to the sensor. For objects with flat and smooth surfaces, a large gripping force is needed to make the method reliable. The method can detect any directional slip without any prior knowledge of the grasped object. In addition, we demonstrate the method provides useful real-time incipient slip feedback in the bottle cap screwing and unscrewing experiment. It can be potentially applied to many manipulation tasks that requires slip detection feedback, such as grasp force control, in hand manipulation, and tool use.







\bibliographystyle{IEEEtran}
\bibliography{Ref}

\end{document}